\documentclass{article}
\usepackage{ijcai22}

\usepackage{times}
\usepackage{soul}
\usepackage{url}
\usepackage[hidelinks]{hyperref}
\usepackage[utf8]{inputenc}
\usepackage[small]{caption}
\usepackage{graphicx}
\usepackage{amsmath}
\usepackage{amssymb}
\usepackage{amsthm}
\usepackage{booktabs}
\usepackage{algorithm}
\usepackage{algorithmic}
\usepackage{subcaption}
\usepackage{cleveref}
\usepackage{booktabs}
\usepackage{comment}
\usepackage{xcolor}
\usepackage{array,multirow}
\usepackage{colortbl}
\usepackage{subcaption}

\DeclareMathOperator*{\argmax}{arg\,max}

\newboolean{include-notes}
\setboolean{include-notes}{true}
\newcommand{\sm}[1]{\ifthenelse{\boolean{include-notes}}
 {{\color{orange}SM: #1}}{}}
 \newcommand{\ready}[1]{\ifthenelse{\boolean{include-notes}}
 {{\color{blue} #1}}{}}
\newcommand{\ff}[1]{\ifthenelse{\boolean{include-notes}}
 {{\color{purple}FF: #1}}{}}
\newcommand{\nt}[1]{\ifthenelse{\boolean{include-notes}}
 {{\color{red}NT: #1}}{}}
 
\newcommand{\pl}{\textbf{PL}}
\newcommand{\fimp}{\textbf{FI}}
\newcommand{\lpm}{\textbf{LPM}}
\newcommand{\statespace}{\mathcal{S}}
\newcommand{\actionspace}{\mathcal{A}}
\newcommand{\nn}{NN}

\newcommand{\newcite}[1]{\citeauthor{#1}~\shortcite{#1}}  
\newcommand{\numpapers}{49} 
\newcommand{\numuserstudies}{12}
\newcommand{\numcandidate}{92}

\definecolor{lpmcolor}{rgb}{0.8706,.921568,.9686274} 
\definecolor{ficolor}{rgb}{0.88627,0.815686,0.941176} 
\definecolor{plcolor}{rgb}{0.9843,0.898,0.83922} 

\title{A Survey of Explainable Reinforcement Learning}

\author{
Stephanie Milani\thanks{Equal contribution}$^1$
\and
Nicholay Topin\footnotemark[1]$^1$\and
Manuela Veloso$^{1}$\And
Fei Fang$^1$
\affiliations
$^1$Carnegie Mellon University}

\begin{document}

\maketitle

\begin{abstract}
Explainable reinforcement learning (XRL) is an emerging subfield of explainable machine learning that has attracted considerable attention in recent years.
The goal of XRL is to elucidate the decision-making process of learning agents in sequential decision-making settings. 
In this survey, we propose a novel taxonomy for organizing the XRL literature that prioritizes the RL setting.
We overview techniques according to this taxonomy. 
We point out gaps in the literature, which we use to motivate and outline a roadmap for future work.
\end{abstract}

\section{Introduction}
In reinforcement learning (RL), an agent learns to take actions in an environment to maximize the accumulated reward through trial and error. 
This framework has been successfully applied to many sequential decision-making problems, like games~\cite{mnih2013playing}, robotics~\cite{ImpRoboticController}, and more.
However, the difficulty of verifying and predicting the behavior of RL agents often hampers their real-world deployment.
This problem is exacerbated when RL is combined with the generalization and representational power of deep neural networks (\nn s).
Without an understanding of how the agent works, it is hard to intervene when necessary or trust that the agent will act reasonably and safely. 

Recently, there has been increasing interest in \textit{explaining} RL models to gain insight into the agent's decision-making process.
The catalyzation of this interest can be traced to the DARPA XAI project~\cite{gunning2017explainable}, which evolved from a narrower focus on supervised learning to broader AI tasks, including RL.
Explainable RL (XRL) offers unique research opportunities: we can apply techniques from explainable supervised learning and also develop methods that account for and utilize the sequential nature of RL problems.

We present a comprehensive survey of XRL literature, organized based on our novel proposed taxonomy. 
Existing surveys on XRL focus on delving deeper into a more limited set of papers~\cite{alex2020explainability,puiutta2020explainable,wells2021explainable} or assessing the need for XRL in different domains instead of summarizing techniques~\cite{zelvelder2021assessing}.
In contrast, we present a novel taxonomy for XRL research and cover a larger and more up-to-date set of papers. 

\begin{figure}[t]
  \centering
  \includegraphics[width = .95\columnwidth]{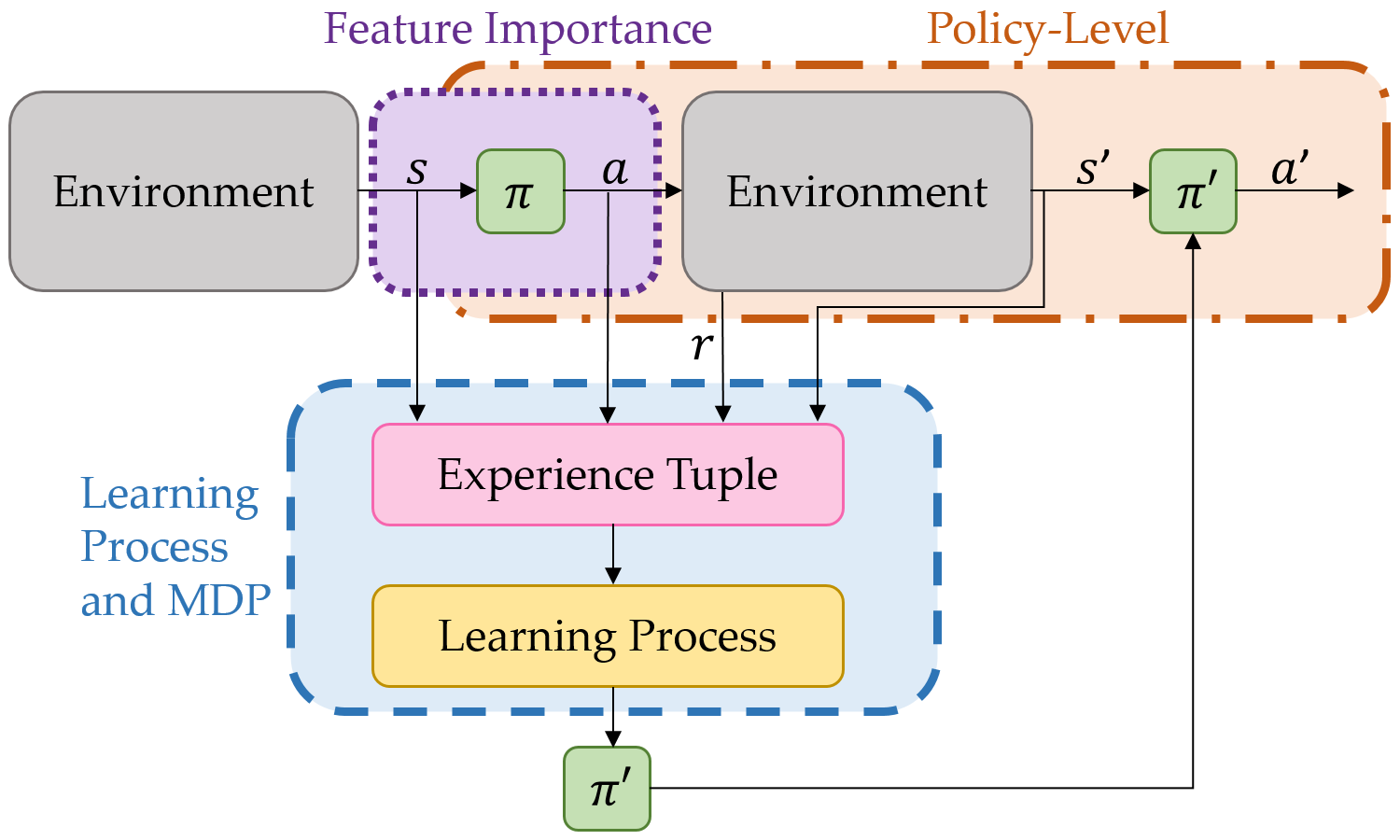}
  \caption{XRL taxonomy and its relationship to the RL process.}
  \label{fig_taxonomy}
\vspace{-10pt}
\end{figure}

\paragraph{Scope.} We focus on papers in which the authors explicitly identify explainability or interpretability as a goal of the work. Since we mainly care about the learning and the behavior of the RL agent,
we omit papers that focus only on explaining the environment in which the agent acts.
We only consider techniques developed for a \textit{single} agent. 
We limit our survey to published work. 
After curating a smaller set of \numcandidate{} candidate papers that satisfy the inclusion criteria, we produce a representative sample of \numpapers{} papers in this area.

Our contributions are summarized as follows: \\
\noindent $\boldsymbol{\cdot}$ \textbf{New taxonomy:} We present a new taxonomy for understanding XRL research. 
We propose to organize techniques according to the part of the RL agent they explain: the immediate context for single actions (termed feature importance or \fimp), influential experiences from training (termed Learning Process and MDP or \lpm), and long-term behavior (termed policy-level or \pl).
~\Cref{fig_taxonomy} depicts the high-level categories of the taxonomy and their relationship to the RL framework. \\
\noindent $\boldsymbol{\cdot}$ \textbf{Comprehensive review:} 
We overview the XRL literature, categorizing representative approaches according to our novel taxonomy.
We compare and contrast the different types of explanation-generation techniques. \\ 
\noindent $\boldsymbol{\cdot}$ \textbf{Future directions:} We highlight gaps in the literature and outline a roadmap for future work, covering: investigating properties of explanations, developing benchmarks for standardized comparisons, and creating RL-specific explanations.

\begin{table*}[t]
\small
\centering
\begin{tabular}{ccl}
\toprule
Category   & Subcategory & Publications \\
\hline
\cellcolor{ficolor} & Learn Intrinsically  & \citeauthor{HEIN2017fuzzyrl}~\shortcite{HEIN2017fuzzyrl,genetic},~\newcite{rodriguez2020optimization},~\newcite{landajuela2021discovering},\\
 
\cellcolor{ficolor} & Interpretable Policy & \newcite{topin2021iterative},~\newcite{zhang2021off} \\\cline{2-3} 
\cellcolor{ficolor} Feature & Interpretable Format & \newcite{Jhunjhunwala_policyextraction_2019},~\newcite{zhang2020interpretable},~\newcite{bewley2021tripletree} \\\cline{2-3}  
\cellcolor{ficolor} Importance & & \newcite{ImpRoboticController},~\newcite{frogger},~\newcite{goel2018unsupervised},\\  
\cellcolor{ficolor} (\textbf{FI}) & Directly Generate & \newcite{greydanus2018visualizing},~\newcite{katia},~\newcite{weitkamp2018visual},\\  
\cellcolor{ficolor} & Explanations & \newcite{annasamy2019towards},~\newcite{huber2019enhancing},~\newcite{mott2019towards},\\ 
\cellcolor{ficolor} &  & 
\newcite{wang2019verbalexplanations},~\newcite{atrey2019exploratory}, \newcite{shi2020selfsupervised},~\newcite{tang2020neuroevolution},\\  
 \cellcolor{ficolor} & & \newcite{itaya2021visual},~\newcite{zhang2021learning} \\  
\hline
\cellcolor{lpmcolor} Learning & Model Domain & \citeauthor{cruz2019memorybasedxrl}~\shortcite{cruz2019memorybasedxrl,cruz2020explainable},~\newcite{chen2020interpretable},~\newcite{madumal2020explainable},\\  
\cellcolor{lpmcolor} Process & Information & \newcite{yau2020did},
~\newcite{lin2021contrastive},~\newcite{volodin2021causeoccam} \\\cline{2-3}  
\cellcolor{lpmcolor} and MDP & Decompose Reward Function & \newcite{anderson2019explaining},~\newcite{beyret2019dot},~\newcite{bica2021learning},~\newcite{guo2021edge} \\\cline{2-3}  
 \cellcolor{lpmcolor} (\textbf{LPM}) & Identify Training Points & \newcite{dao2018deep},~\newcite{gottesman2020interpretable} \\ \cline{2-3}  
\hline 
\cellcolor{plcolor}& Summarize Using Transitions & \newcite{amir2018highlights},~\newcite{huang2018establishing},~\newcite{lage2019exploring}\\\cline{2-3}  
\cellcolor{plcolor} Policy-Level & Convert RNN & \newcite{koul2018learning},~\newcite{danesh2021re},~\newcite{hasanbeig2021deepsynth} \\\cline{2-3}  
\cellcolor{plcolor} (\textbf{PL}) & Extract Clusters or Abstract States& \newcite{cluster_drl},~\newcite{topin2019policylevel},~\newcite{sreedharan2020tldr} \\
\hline
\end{tabular}
\caption{Taxonomy and corresponding representative publications in XRL. Category color corresponds to color used in~\Cref{fig_taxonomy}.}
\label{tab:taxonomy_pubs}
\end{table*}

\section{Background and Preliminaries}
\label{sec:rl_background}

\subsection{Reinforcement Learning}
In RL, an agent learns how to behave in an environment to maximize a reward signal~\cite{sutton2018reinforcement}. 
The environment is defined by a Markov decision process (MDP) $M = ( \statespace, \actionspace, T, R, \gamma )$, where $\statespace$ is the state space, $\actionspace$ is the action space, $T : \statespace \times \actionspace \times \statespace \rightarrow [0, 1]$ is the state-transition function dictating the environment's transition dynamics, $R : \statespace \times \actionspace \times \statespace \rightarrow \mathbb{R}$ is the reward function, and $\gamma \in (0,1]$ is the scalar discount factor that governs the importance of future rewards. 
In RL, the \textit{model} (i.e., $R$ and $T)$ is unknown.
At each time step $t$, an agent observes a state $s_t$ and chooses an action $a_t$ according to its policy $\pi$.
The agent executes $a_t$, receives reward $r_t$ according to $R$, and observes a new state $s_{t+1}'$ according to $T$.
The goal of RL is to find an optimal policy $\pi^*$, specifying which action to take in each state, which results in the highest expected discounted future return: $\pi^* = \argmax_{\pi} Q^\pi (s, a)$, where $Q^\pi(s,a)$ is the \textit{action-value} function.

Large or continuous state spaces prevent exactly storing $Q^\pi$ or $\pi$.
Instead, many RL methods leverage function approximators: most commonly, \nn s.
The input to these networks is often features describing the state.
Agents that use a non-interpretable function approximator require additional steps for a person to understand. 

\subsection{Explainable Machine Learning}
Most work in explainable machine learning focuses on supervised learning, which learns from a labeled data set to predict a label $y$ given input data $x$.
These methods are typically grouped into two categories: intrinsic and post-hoc.
\textit{Intrinsic} interpretability refers to the direct construction of understandable models, such as small decision trees (DTs)~\cite{lipton2018mythos}.
\textit{Post-hoc} interpretability refers to creating an additional model to explain an existing one.
Explanations can also be \textit{local} or \textit{global}.
Local explanations explain the prediction for a specific data point; global explanations holistically view the machine learning model and its predictions. 

We focus on a few key metrics for evaluating explanations.
We refer an interested reader to a more comprehensive overview of metrics~\cite{molnar2019}. 
\textit{Fidelity} measures how faithful the model explanation is to the true prediction.
\textit{Performance} refers to the standard evaluation metric for the task.
\textit{Relevancy} refers to the ability of the explanation to produce insight for a particular audience in a domain~\cite{murdoch2019definitions}.
\textit{Cognitive load} refers to the cognitive resources required for performing a mental task~\cite{abdul2020cogam}.
Relevancy and cognitive load are challenging to operationalize and quantitatively measure, but assessing them is invaluable for producing useful and actionable explanations.

\section{XRL Taxonomy}
\label{sec:taxonomy}
We propose a novel taxonomy that prioritizes the RL setting.
We primarily distinguish between the surveyed literature according to the central goals of explanations for RL.
~\Cref{fig_taxonomy} depicts how these categorizations correspond to parts of the RL framework.
\textit{Feature importance} (\fimp) explanations identify the features that affect an agent’s action choice $a_t$ for the input state $s_t$.
\textit{Learning process and MDP} (\lpm) explanations show the past experiences or the components of the MDP that led to the current behavior.
\textit{Policy-level} (\pl) explanations illustrate the long-term behavior of the agent.

These explanations communicate different aspects of the RL agent.
\fimp~explanations provide an action-level look at the agent's behavior: for each action, one can query for the immediate context that was critical for making that decision.
\lpm~explanations provide additional information about the effects of the training process or the MDP.
Some \lpm~explanations enable people to understand the influential experiences that led to the agent's current behavior. 
Others describe how the agent acts in terms of other MDP components, like reward.
\pl~explanations present summaries of long-term behavior through abstraction or representative examples.
They are critical for understanding an agent's behavior to evaluate its overall competency.
See~\Cref{tab:taxonomy_pubs} for the taxonomy and representative XRL publications. 

\section{Feature Importance (FI)} 
\label{sec:fi}
\begin{figure}
    \centering
    \begin{subfigure}[b]{0.13\textwidth}
        \includegraphics[width=1.0\textwidth]{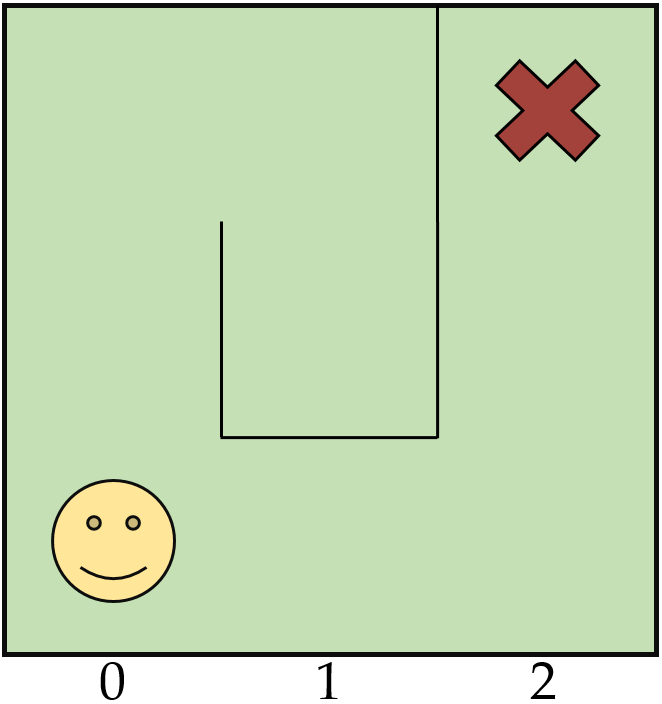}
        \caption{}
    \end{subfigure}
    \hfill
    \begin{subfigure}[b]{0.13\textwidth}
        \includegraphics[width=1.0\textwidth]{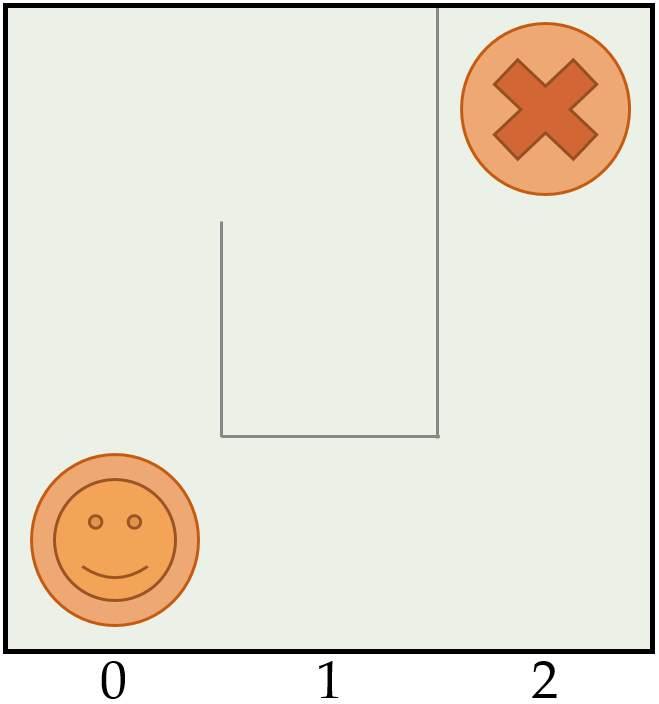}
        \caption{}
    \end{subfigure}
    \hfill
    \begin{subfigure}[b]{0.08\textwidth}
        \includegraphics[width=1.0\textwidth]{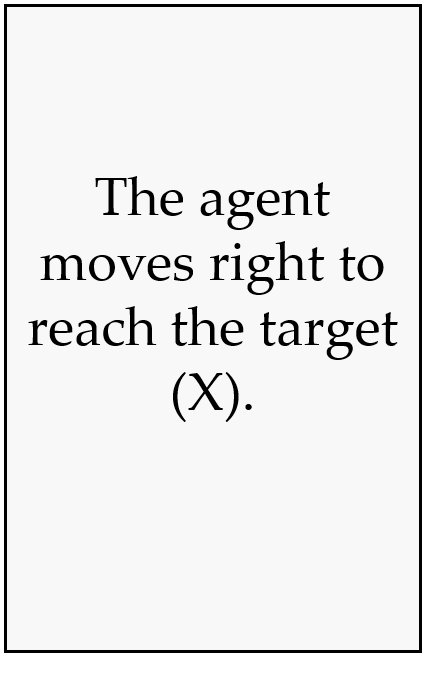}
        \caption{}
    \end{subfigure}
    \hfill
    \begin{subfigure}[b]{0.10\textwidth}
        \includegraphics[width=1.0\textwidth]{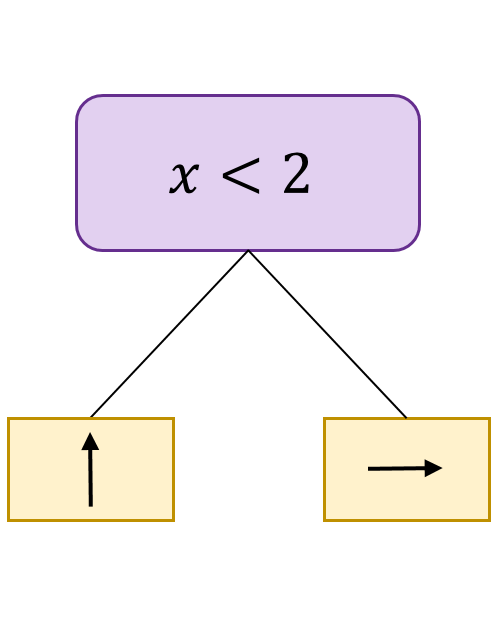}
        \caption{}
    \end{subfigure}
    \caption{Example object saliency map (b), natural language explanation (c), and DT policy (d), given domain state (a).} 
  \label{fig_fi_example_explanations}
\end{figure}

Many \fimp~techniques extend those from the explainable supervised learning literature.
Viewed in the supervised learning framework, the input data is the state, and the label is the action.
Some \fimp~techniques directly learn an intrinsically interpretable policy (\Cref{rw_peraction_inherent}); others produce an intrinsically interpretable policy through post-hoc conversion of a non-interpretable one (\Cref{rw_peraction_convert}).
Other methods do not learn an interpretable policy but instead generate action explanations in various forms, e.g., natural language (Section~\ref{rw_peraction_generate}). 
~\Cref{fig_fi_example_explanations} depicts an example of different types of explanations.

\subsection{Learn an Intrinsically Interpretable Policy}
\label{rw_peraction_inherent}
We overview methods that directly learn intrinsically interpretable policies.
We distinguish between those that learn DT policies and those that learn other policy representations.

DTs are not differentiable, so~\newcite{rodriguez2020optimization} propose to learn a soft DT, in which sigmoid activation functions replace the Boolean decisions in classic DTs.
This tree can then be discretized (at the cost of performance) to form a DT-form policy approximation. 
In contrast, CUSTARD~\cite{topin2021iterative} constructs an augmented MDP where the action set contains actions for constructing a DT, i.e., choosing a feature to branch a node. By training the agent with the augmented MDP, the agent directly learns a DT policy for the original environment while still using \nn s during training.
However, DTs have drawbacks: they produce only axis-parallel partitions, cannot handle approximate reasoning, and are non-differentiable.
Alternative representations mitigate these issues:
trees with algebraic expressions represented by nodes~\cite{genetic,landajuela2021discovering} can represent complex functions, fuzzy controllers~\cite{HEIN2017fuzzyrl} enable approximate reasoning, and certain logic formulations~\cite{zhang2021off} permit differentiable learning.

These techniques enjoy the benefit of producing an intrinsically interpretable policy. 
However, some representations risk more initial cognitive load from a user to understand, so they are more applicable when their long-term benefit outweighs the cost of the initial scaffolding for understanding explanations~\cite{dodge2021no}.

\subsection{Convert Policy to Interpretable Format}
\label{rw_peraction_convert}
Some methods approximate an existing, performant \nn~policy $\pi$ with an interpretable \textit{surrogate} model $\hat{\pi}$.
To learn $\hat{\pi}$, we can use \textit{imitation learning} (IL)~\cite{abbeel2004apprenticeship} --- also known as \textit{learning from demonstration}~\cite{argall2009survey} --- where the RL policy $\pi$ is the \textit{expert}, and the IL policy $\hat{\pi}$ is the \textit{learner}.
The problem becomes a classification problem: the learner aims to correctly classify which action (label) to take from a state, given examples from the expert. 

These techniques augment the IL process, expand the data used for training, or propose a new type of surrogate model. 
VIPER~\cite{bastani2018verifiable} trains a sequence of DTs on data points sampled based on how ``critical'' they are, measured by $V^{\pi}(s) - \min_{a \in \actionspace} Q^\pi (s,a)$.
PIRL~\cite{verma2018programmatically} locally searches over interpretable programmatic policies by minimizing the distance between $\pi$'s outputs and $\hat{\pi}$'s outputs on heuristically-chosen states.
~\newcite{Jhunjhunwala_policyextraction_2019} distills deep RL policies into Q-value-based trees with nodes that encode more complex boundary expressions.
~\newcite{zhang2020interpretable} derive $\hat{\pi}$ from $\pi$ with genetic programming.
TripleTree~\cite{bewley2021tripletree} expands the information used in the DT-learning algorithm.
\newcite{liu2018toward} introduces DTs with linear models at the leaves and proposes to use stochastic gradient descent to update their values.

An IL policy is necessarily based only on the states and actions encountered during the example executions from the expert, so it may not be robust to novel situations.
As robust IL techniques advance and become incorporated with RL, we expect even more developments in post-hoc XRL.
Furthermore, these papers show that it is possible to achieve reasonable performance and fidelity, but the explanation is still the approximated model, not the original policy.

\subsection{Directly Generate Explanation}
\label{rw_peraction_generate}
Action explanations generated from a non-interpretable policy most commonly take the forms of natural language and saliency maps.
Natural-language explanations linguistically describe the agent's decision-making process. 
Saliency maps are images that topographically highlight regions that are important for an agent's decision. 
We also discuss explanations that cannot be categorized as one of the previous types. 

\paragraph{Natural Language.}
We distinguish natural-language explanations by their means of production. 
Some work leverages templates for the agent to fill in~\cite{ImpRoboticController}; other work allows the agent to generate free-form explanations.
The former type typically enjoys more clarity but requires the explanation to adhere to a specified template. 
Furthermore, there is some question about the validity of these explanations. 
~\newcite{frogger} and~\newcite{wang2019verbalexplanations} leverage human-provided action-explanation pairs to \textit{rationalize} agent behavior. 
We caution against post-hoc rationalizations that produce plausible but not valid explanations: they may increase agent trust, but the trust is unfounded without grounding to the agent's decision-making process.

\begin{table}[t]
\small
    \centering
    \begin{tabular}{cl}
    \toprule 
    Type     &  Publications\\
    \midrule
      & \newcite{cluster_drl},~\newcite{weitkamp2018visual},\\
    Gradient & \newcite{huber2019enhancing} \\
    \hline
    Perturbation & \newcite{greydanus2018visualizing},~\newcite{rupprecht2020finding}\\ 
    \hline
     & \newcite{katia},~\newcite{goel2018unsupervised}, \\
    Object & \newcite{annasamy2019towards}\\
    \hline 
    & \newcite{mott2019towards},~\newcite{shi2020selfsupervised}, \\
    Attention & \newcite{tang2020neuroevolution},~\newcite{itaya2021visual}, \\
    & \newcite{zhang2021learning} \\
    \bottomrule
    \end{tabular}
    \caption{Saliency maps and their associated categorization.}
    \label{tab:saliency}
\end{table}

\paragraph{Saliency Maps.} 
We categorize saliency map techniques using the decomposition in~\newcite{atrey2019exploratory} (see~\Cref{tab:saliency}). 
For example~\newcite{greydanus2018visualizing} construct saliency maps by perturbing the input image describing the state with Gaussian blur, then measuring changes in the policy after removing information from an area. 
For more details, we refer the reader to~\newcite{atrey2019exploratory}.
Saliency is viewed as insufficient for explaining RL, as the conclusions drawn from it are highly subjective~\cite{atrey2019exploratory}.
Indeed, the majority of claims in the surveyed papers either propose an \textit{ad hoc} explanation of agent behavior after observing saliency or develop an \textit{a priori} explanation of behavior evaluated using saliency.
In real-world settings, we desire more reliable explanations.

\paragraph{Other.}
There exist other forms of explanation, including counterfactual states~\cite{olson2021counterfactual}, which illustrate the minimal change needed to induce a different action selection $a_t'$.
This area is ripe for exploration, especially given the recent interest in multimodal machine learning~\cite{baltruvsaitis2018multimodal}. 
In this setting, we may want custom local explanations that reflect different modalities.

\section{Learning Process and MDP (\lpm)} 
\label{rw_trainingimpact}
\lpm~explanations reveal the influence of other parts of the MDP (e.g., $R$ and $T$) or the interactions with the MDP during learning on the agent's behavior.
Most commonly, the agent reveals its learned transition dynamics $\hat{T}$ (\Cref{rw_trainingimpact_tf}). 
Alternatively, techniques decompose the reward function to produce explanations in terms of the agent's objectives (\Cref{rw_trainingimpact_rf}).
Other techniques determine training points that influenced the learned policy (\Cref{rw_trainingpoints}).

\subsection{Model Domain Information} 
\label{rw_trainingimpact_tf}
Approaches in this category learn models $\hat{T}$ that approximate $T$ and generate explanations based on $\hat{T}$.
They differ in terms of the questions these generated explanations answer.

To answer ``what-if'' questions,~\newcite{rupprecht2020finding} learn a generative model of the environment, which enables the probing of agent behavior in novel states created by an optimization scheme that induces certain actions in agents.
For example, one can optimize for states where the agent expects to perform poorly, then visualize such states to observe the agent's behavior.
To contrast between a selected action $a_t$ and alternative ones $a_t'$,~\citeauthor{cruz2019memorybasedxrl}~\shortcite{cruz2019memorybasedxrl,cruz2020explainable} use episodic memory to explain decisions with success probability, i.e., ``$a_t$ gives an 85\% probability of task success compared to 38 \% of $a_t'$''. 
~\newcite{lin2021contrastive} provide action preference explanations in terms of human-provided features.
~\newcite{yau2020did} provide explanations in terms of intended outcome: the agent projects predicted future trajectories from its current state and proposed action. 
To answer ``why?'' and ``why not?'' questions,~\newcite{madumal2020explainable} leverage an action influence graph, which is a structural causal model (with pre-specified, potentially learned~\cite{volodin2021causeoccam} structure) that encodes causal action-based relationships between state features of interest.
To understand correlative context for failure cases,~\newcite{chen2020interpretable} learn a sequential latent environment model that can be used to generate a semantic mask to understand the surrounding context.

Explanations can address a plethora of potential questions a user may have. 
Clearly defining these questions and the ability of the techniques to address them is important for users to understand an explanation's capabilities.
For example, we do not want users to assume causal relationships when the explanation only provides correlations.
    
\subsection{Decompose Reward Function}
\label{rw_trainingimpact_rf}
By conveying aspects of the reward that contributed to the agent's behavior, an explanation provides information in terms of the agent's objectives. 
Most work in this area focuses on presenting interpretable views of $Q^\pi(s,a)$.

Before the agent starts learning, the explanation designer can use reward decomposition charts (RDCs)~\cite{juozapaitis2019explainable,anderson2019explaining} to split $R(s,a)$ into a set of additive terms with semantic meaning: $R(s,a) = \sum_{c \in C} R_c(s,a)$, where $C$ is the set of reward types.
$Q(s,a)$ is similarly decomposed as $Q^\pi(s,a) = \sum_{c \in C} Q^\pi_c(s,a)$.
With this decomposition, one can inspect actions in terms of trade-offs between the types. 
~\newcite{bica2021learning} model an expert's reward function in terms of preferences over proposed alternative outcomes to produce explanations of the trade-offs associated with actions. 
~\newcite{beyret2019dot} decompose the agent's task into subtasks to visualize per-subtask Q-values as a heatmap over the environment.
EDGE~\cite{guo2021edge} learns to identify which timesteps contributed to the final reward to identify policy errors and weaknesses.

Reward-based explanations may help laypeople understand how reward-based objectives influence the decision-making process of an RL policy~\cite{anderson2019explaining}.
However, we require further explanatory steps to produce actionable explanations. 
For example, RDCs could be combined with an approach from~\Cref{rw_trainingpoints} to illustrate which experiences impact each of the $Q^\pi_c$ estimates. 
Ideally, this combination also reveals the experiences to remove to affect the estimates of these different components.
    
\subsection{Identify Training Points}
\label{rw_trainingpoints}
These techniques identify important training points, or transition tuples, that influence the agent's learned behavior.
Influence refers to a change in some function value (e.g., $Q(s,a)$).

\newcite{dao2018deep} select and record states with associated importance weights that correspond to different decisions in a learned policy. 
Equipped with this knowledge, one can understand the most influential transition tuples for a policy's individual decisions. 
However, it is unclear how to amend the training process to remove undesirable behaviors based on this information alone. 
~\newcite{gottesman2020interpretable} identify training points that are \textit{influential} for the estimation of Q-values (i.e., whose removal causes a large change in Q-value estimates for initial states). 
This technique enables users to determine if there is sufficient data for learning the desired behavior and what type of data is lacking. 

\section{Policy-Level (\pl)} 
\label{rw_policylevel}
A less-studied but important category of explanations is those which describe an agent's longer-term behavior. 
We discuss three sub-categories: summarizing behavior in the context of experiences encountered by the agent during training (\Cref{pl_transitions}), converting recurrent neural networks (RNNs) to interpretable forms (\Cref{rw_policylevel_fsm}), and summarizing behavior according to similar states (\Cref{rw_policylevel_mc}).

\subsection{Summarize Using Transitions}
\label{pl_transitions}
We can explain an agent's behavior in terms of the experiences encountered by the agent during training.
With these explanations, human operators know which experiences influence the agent's behavior, providing more information to adjust the behavior if desired. 
These methods are distinct from those described in~\Cref{rw_trainingpoints} because the goal is to describe the agent's \textit{long-term} behavior based on influential \textit{trajectories}, not individual transition tuples.
These trajectories are combined to create a summary of agent behavior.

These methods use different selection criteria for choosing the trajectories or transition tuples.
~\newcite{amir2018highlights} select trajectories based on the state importance $I(s) = \max_{a} Q^\pi(s,a) - \min_a Q^\pi(s,a)$ and diversity, which is considered by identifying the most similar state $s_s$ to a state $s$ currently included in the summary and comparing $I(s_s)$ with $I(s)$. 
The more important state and its corresponding trajectory is used in the summary.
Similarly,~\newcite{huang2018establishing} select states where the chosen action has a much higher Q-value than another action.
To investigate the effects of different summary-extraction models on policy reconstruction,~\newcite{lage2019exploring} compare approaches for choosing subsets of $s,a$ pairs that characterize the agent's behavior.
They motivate personalizing user models for summarization, underscoring the importance of relevancy as an evaluation metric.

These metrics for importance tend to highlight \textit{extreme} policy behavior: states in which the agent risks performing very poorly compared to its best action. 
This property is useful for high-stakes, real-world scenarios.
However, these techniques are not helpful for understanding how these behaviors fit together.
This problem is exacerbated when scaling the environment complexity, as the trajectories may be so widely temporally spaced that the interpretation is not useful.
    
\subsection{Convert RNN to Interpretable Format}  
\label{rw_policylevel_fsm}
RNNs commonly represent policies because their internal memory encodes features of the observation history critical for decision making.
One way to understand these structures is through learning more compact memory representations by converting the RNN to a finite state machine (FSM)~\cite{koul2018learning}.
However, this minimization process can merge semantically-distinct states, obfuscating a deeper understanding of the FSM. 
As a result,~\newcite{danesh2021re} propose a more interpretable reduction technique to preserve key decision points of the policy. 
This family of techniques only applies in cases where the agent contains a RNN. 
In contrast,~\newcite{hasanbeig2021deepsynth} construct a deterministic finite automaton to capture these sequential dependencies in an intrepretable representation without reliance on an RNN policy.

Completely removing the RNN from the agent permits a more interpretable structure.
However, the agent is changed through the minimization process, yet it is treated as equivalent to the original agent.
In other words, the agent acts \textit{like} it is following the resulting FSM; however, it is not internally behaving this way.

\subsection{Extract Clusters or Abstract States}
\label{rw_policylevel_mc}
Sometimes it is invaluable to understand the overall behavior of the agent in terms of how it will act when it encounters similar states.
One advantage of this approach is that people only need to understand the similarity metric for aggregating states and the model for summarizing the policy.
XRL methods that produce summarizations using these clustered or abstract states are similar to those for generating \textit{abstractions} for RL.
We refer an interested reader to recent work~\cite{abel2020theory} on abstraction-generation methods.
  
TLdR~\cite{sreedharan2020tldr} identifies landmarks, which are defined as propositional formulas that must be satisfied for an agent to complete a goal. 
The possible transitions between landmarks are displayed as a graph depicting how an agent would reach a goal state from the starting state.
Abstract policy graphs~\cite{topin2019policylevel} are Markov chains of abstract states that summarize a policy in terms of expected future transitions. 
~\newcite{cluster_drl} do not directly construct graphs; instead, states are embedded into a space, where the agent treats states located close to one another in this space similarly.
A human operator can identify clusters and relationships between these clusters through manual inspection.
  
These techniques rely on different assumptions, such as the binarization of features~\cite{topin2019policylevel}. 
Future work could investigate relaxing assumptions to produce more general techniques.
Furthermore, we may desire different semantic graph structures, such as hierarchical graphs, where the high-level graph describes the subtask transitions and the lower-level graphs describe the lower-level control policies.

\section{Discussion}
\label{sec:discussion}
We compare and contrast \fimp, \lpm, and \pl{} explanations, focusing specifically on what these techniques do and do not enable. 
With these comparisons, we aim to demonstrate the trade-offs between the different types of explanations.

\fimp~explanations communicate the local behavior, or individual action choice, of an agent in a particular situation. 
Some techniques guarantee that the explanations capture the information used in the agent's decisions, but others do not.
For example, techniques that directly produce DT policies necessarily elucidate the features and corresponding values that influence the agent's decision.
In contrast, techniques that generate post-hoc \textit{rationalizations}, like some natural language explanations, produce explanations based on a human's decision-making process when solving a task.
Furthermore, \fimp{} explanations do not provide global behavioral summaries that convey subtask or plan information: they only explain individual action choices.
\pl~explanations, however, provide this information by producing summaries using transitions, model approximations, or similar states.

\lpm~explanations enable understanding of the influence of the MDP or training process on the agent.
However, they generally require extra information or learning during training, so we cannot generate them post-hoc. 
For example, RDCs require the explanation designer to use domain information to decompose the reward function before any learning occurs.
In contrast, many \fimp~and \pl~explanations do not require additional information and/or learning, so we can generate them post-hoc. 

\pl~explanations permit the understanding of long-term behavior.
Because these explanations are summaries, they necessarily exclude details for the sake of interpretability. 
For example, techniques from~\Cref{rw_policylevel_mc} exclude information in the state aggregation techniques by definition of abstraction.
In contrast, intrinsically-interpretable \fimp~explanations explicitly include all of the information available about how the agent makes choices.

\section{A Roadmap for XRL}
\label{sec:future_work}
Here we synthesize the gaps in the XRL literature to discuss opportunities for future research.
We focus on the following directions: investigating the properties of explanations for RL, standardizing evaluation through benchmarks, and focusing on techniques that leverage the unique facets of the RL problem to provide correct and coherent explanations.

\subsection{Properties}
\label{sec:properties}
We believe it is important to investigate properties of explanations for RL and to standardize these evaluations through operationalized metrics. 
Some metrics, such as cognitive load and relevancy, necessitate user studies to properly evaluate; however, only \numuserstudies{} of the \numpapers{} surveyed papers include user studies.
Unfortunately, some studies give an incomplete picture of the utility of the proposed explanations.
For example, it is not enough to evaluate whether people \textit{believe} explanations are useful; we must also evaluate whether they \textit{are} useful and accurate in explaining the agent's behavior.
 
When proposing a new form of explanation, we recommend performing a user study to evaluate if the explanation is practical and understandable. 
Measuring comprehensibility tells us whether the explanation is understandable at all, whereas measuring the associated cognitive load provides insight into how challenging the explanation is to interpret.
In these studies, we also suggest accounting for influential characteristics of the users, such as domain expertise, to evaluate the \textit{relevance} of the explanations~\cite{ehsan2020human}.

We caution against substituting explanations with ad-hoc rationalizations.
Instead, explanations ought to reflect the agent's underlying decision-making process, rather than a post-hoc justification of behavior based on subjective interpretation~\cite{atrey2019exploratory}.
With these more accurate explanations, we can not only increase user trust in the system but also ensure that this trust is well-founded.

Finally, we encourage researchers to provide concrete descriptions of what their methods enable.
For example, it can be misleading to provide explanations phrased as causal relationships for techniques that only explain correlations.
Transparency surrounding the affordances of explanations is critical to prevent user misunderstanding.

\subsection{Benchmarks}
Domains used in the literature to evaluate explanations vary widely, from toy control~\cite{zhang2021off} to games~\cite{olson2021counterfactual} to real healthcare data~\cite{bica2021learning}. 
We believe it is critical to evaluate the quality of explanations using benchmarks with standardized environments and metrics.
Benchmarks, and competitions based off of these benchmarks~\cite{cobbe2020leveraging}, can propel research forward by providing appropriate scope and focus to an important problem.
To that end, we propose two types of benchmarks for XRL research: \textit{type-based} and \textit{objective-based}.

Type-based benchmarks aim to produce high-quality explanations of a \textit{certain} type, e.g., DT policies, which are evaluated in a standardized set of environments with common metrics. 
If possible, these environments should permit evaluation against some ground-truth.
For example, when comparing DTs, it should be possible to represent $\pi^*$ with a DT in (at least) the simple environments (e.g., Cartpole) to ensure that the techniques can reliably learn the optimal DT policy. 
In an objective-based benchmark, the goal is to produce high-quality explanations of \textit{any} type for a particular task.
The goal is to produce explanations that score the highest on a set of evaluation metrics, which can include comprehensibility and predictability of the system by users.

In both types of benchmarks, we require human-in-the-loop evaluations to properly compare techniques against one another.
To our knowledge, only one benchmark provides a principled means of doing such an evaluation~\cite{shah2021minerl}; however, it is done in the context of learning from hard-to-define reward functions, not interpretability.
As a result, another critical research direction is developing a principled approach for human-in-the-loop evaluations~\cite{dodge2021no} of XRL techniques.

\subsection{RL-Specific Explanations}
As shown in~\Cref{tab:taxonomy_pubs}, a large portion of XRL techniques stem from the supervised-learning literature. 
In particular, \fimp{} techniques mirror the standard interpretable classification setup by providing local explanations of single actions (the label) in the context of the immediate state (the input data to be classified).
Although these explanations are useful for elucidating the immediate context of local decisions, we can leverage important components of the RL problem to produce more actionable and informative explanations. 
As a result, we believe a fruitful direction of research is one which leverages these critical components, such as reward to communicate agent objectives (\Cref{rw_trainingimpact_rf}), causal models of the agent's interactions with the environment to understand the agent's impact (\Cref{rw_trainingimpact_tf}), and long-term behavior to depict how an agent will behave over time (\Cref{rw_policylevel}). 

\section{Conclusion}
We presented a comprehensive summary of XRL techniques in the context of our novel taxonomy.
We compared representative approaches of each type and discussed their applicability and critiques.
We identified gaps in the current literature and concluded by suggesting three major avenues of research to drive the field forward.

\section*{Acknowledgments}
This research was sponsored by the U.S. Army Combat Capabilities Development Command Army Research Laboratory and was accomplished under Cooperative Agreement Number W911NF-13-2-0045 (ARL Cyber Security CRA). 
The views and conclusions contained in this document are those of the authors and should not be interpreted as representing the official policies, either expressed or implied, of the Combat Capabilities Development Command Army Research Laboratory or the U.S. Government. 
The U.S. Government is authorized to reproduce and distribute reprints for Government purposes notwithstanding any copyright notation here on.
This material is based upon work supported by the Department of Defense (DoD) through the National Defense Science \& Engineering Graduate (NDSEG) Fellowship Program. 
Any opinions, findings and conclusions or recommendations expressed in this material are those of the author(s) and do not necessarily reflect the views of the Department of Defense, Defense Advanced Research Projects Agency or the Air Force Research Laboratory.

\bibliographystyle{named}
\bibliography{ijcai}

\begin{thebibliography}{}

\bibitem[\protect\citeauthoryear{Abbeel and
  Ng}{2004}]{abbeel2004apprenticeship}
P~Abbeel and AY~Ng.
\newblock Apprenticeship learning via inverse reinforcement learning.
\newblock In {\em ICML}, 2004.

\bibitem[\protect\citeauthoryear{Abdul \bgroup \em et al.\egroup
  }{2020}]{abdul2020cogam}
A~Abdul, C~von~der Weth, et~al.
\newblock Cogam: measuring and moderating cognitive load in machine learning
  model explanations.
\newblock In {\em CHI}, 2020.

\bibitem[\protect\citeauthoryear{Abel and Littman}{2020}]{abel2020theory}
D~Abel and M~Littman.
\newblock A theory of abstraction in reinforcement learning.
\newblock {\em Brown Univ.}, 2020.

\bibitem[\protect\citeauthoryear{Amir and Amir}{2018}]{amir2018highlights}
D~Amir and O~Amir.
\newblock Highlights: Summarizing agent behavior to people.
\newblock In {\em AAMAS}, 2018.

\bibitem[\protect\citeauthoryear{Anderson \bgroup \em et al.\egroup
  }{2019}]{anderson2019explaining}
A~Anderson, J~Dodge, et~al.
\newblock Explaining reinforcement learning to mere mortals: An empirical
  study.
\newblock In {\em IJCAI}, 2019.

\bibitem[\protect\citeauthoryear{Annasamy and
  Sycara}{2019}]{annasamy2019towards}
RM~Annasamy and K~Sycara.
\newblock Towards better interpretability in deep q-networks.
\newblock In {\em AAAI}, 2019.

\bibitem[\protect\citeauthoryear{Argall \bgroup \em et al.\egroup
  }{2009}]{argall2009survey}
BD~Argall, S~Chernova, et~al.
\newblock A survey of robot learning from demonstration.
\newblock {\em Robot. Auton. Syst.}, 2009.

\bibitem[\protect\citeauthoryear{Atrey \bgroup \em et al.\egroup
  }{2020}]{atrey2019exploratory}
A~Atrey, K~Clary, and D~Jensen.
\newblock Exploratory not explanatory: Counterfactual analysis of saliency maps
  for deep rl.
\newblock In {\em ICLR}, 2020.

\bibitem[\protect\citeauthoryear{Baltru{\v{s}}aitis \bgroup \em et al.\egroup
  }{2018}]{baltruvsaitis2018multimodal}
T~Baltru{\v{s}}aitis, C~Ahuja, and LP~Morency.
\newblock Multimodal machine learning: A survey and taxonomy.
\newblock {\em TPAMI}, 2018.

\bibitem[\protect\citeauthoryear{Bastani \bgroup \em et al.\egroup
  }{2018}]{bastani2018verifiable}
O~Bastani, Y~Pu, and A~Solar-Lezama.
\newblock Verifiable reinforcement learning via policy extraction.
\newblock In {\em NeurIPS}, 2018.

\bibitem[\protect\citeauthoryear{Bewley and Lawry}{2021}]{bewley2021tripletree}
T~Bewley and J~Lawry.
\newblock Tripletree: A versatile interpretable representation of black box
  agents and their environments.
\newblock In {\em AAAI}, 2021.

\bibitem[\protect\citeauthoryear{Beyret \bgroup \em et al.\egroup
  }{2019}]{beyret2019dot}
B~Beyret, A~Shafti, and AA~Faisal.
\newblock Dot-to-dot: Explainable hierarchical reinforcement learning for
  robotic manipulation.
\newblock In {\em IROS}, 2019.

\bibitem[\protect\citeauthoryear{Bica \bgroup \em et al.\egroup
  }{2021}]{bica2021learning}
I~Bica, D~Jarrett, et~al.
\newblock Learning ``what-if" explanations for sequential decision-making.
\newblock In {\em ICLR}, 2021.

\bibitem[\protect\citeauthoryear{Chen \bgroup \em et al.\egroup
  }{2020}]{chen2020interpretable}
J~Chen, S~Eben Li, and M~Tomizuka.
\newblock Interpretable end-to-end urban autonomous driving with latent deep
  reinforcement learning.
\newblock {\em T-ITS}, 2020.

\bibitem[\protect\citeauthoryear{Cobbe \bgroup \em et al.\egroup
  }{2020}]{cobbe2020leveraging}
K~Cobbe, C~Hesse, et~al.
\newblock Leveraging procedural generation to benchmark reinforcement learning.
\newblock In {\em ICML}, 2020.

\bibitem[\protect\citeauthoryear{Cruz \bgroup \em et al.\egroup
  }{2019}]{cruz2019memorybasedxrl}
F~Cruz, R~Dazeley, and P~Vamplew.
\newblock Memory-based explainable reinforcement learning.
\newblock {\em AI*IA}, 2019.

\bibitem[\protect\citeauthoryear{Cruz \bgroup \em et al.\egroup
  }{2021}]{cruz2020explainable}
F~Cruz, R~Dazeley, and P~Vamplew.
\newblock Explainable robotic systems: Understanding goal-driven actions in a
  reinforcement learning scenario.
\newblock {\em Neural. Comput. Appl.}, 2021.

\bibitem[\protect\citeauthoryear{Danesh \bgroup \em et al.\egroup
  }{2021}]{danesh2021re}
MH~Danesh, A~Koul, et~al.
\newblock Re-understanding finite-state representations of recurrent policy
  networks.
\newblock In {\em ICML}, 2021.

\bibitem[\protect\citeauthoryear{Dao \bgroup \em et al.\egroup
  }{2018}]{dao2018deep}
G~Dao, I~Mishra, and M~Lee.
\newblock Deep reinforcement learning monitor for snapshot recording.
\newblock In {\em ICMLA}, 2018.

\bibitem[\protect\citeauthoryear{Dodge \bgroup \em et al.\egroup
  }{2021}]{dodge2021no}
J~Dodge, A~Anderson, et~al.
\newblock From “no clear winner” to an effective explainable artificial
  intelligence process: An empirical journey.
\newblock {\em Appl. AI Letters}, 2021.

\bibitem[\protect\citeauthoryear{Ehsan and Riedl}{2020}]{ehsan2020human}
U~Ehsan and MO~Riedl.
\newblock Human-centered explainable ai: Towards a reflective sociotechnical
  approach.
\newblock In {\em HCII}, 2020.

\bibitem[\protect\citeauthoryear{Ehsan \bgroup \em et al.\egroup
  }{2018}]{frogger}
U~Ehsan, B~Harrison, et~al.
\newblock Rationalization: A neural machine translation approach to generating
  natural language explanations.
\newblock {\em AIES}, 2018.

\bibitem[\protect\citeauthoryear{Goel \bgroup \em et al.\egroup
  }{2018}]{goel2018unsupervised}
V~Goel, J~Weng, and P~Poupart.
\newblock Unsupervised video object segmentation for deep reinforcement
  learning.
\newblock In {\em NeurIPS}, 2018.

\bibitem[\protect\citeauthoryear{Gottesman \bgroup \em et al.\egroup
  }{2020}]{gottesman2020interpretable}
O~Gottesman, J~Futoma, et~al.
\newblock Interpretable off-policy evaluation in reinforcement learning by
  highlighting influential transitions.
\newblock In {\em ICML}, 2020.

\bibitem[\protect\citeauthoryear{Greydanus \bgroup \em et al.\egroup
  }{2018}]{greydanus2018visualizing}
S~Greydanus, A~Koul, et~al.
\newblock Visualizing and understanding atari agents.
\newblock In {\em ICML}, 2018.

\bibitem[\protect\citeauthoryear{Gunning}{2017}]{gunning2017explainable}
D~Gunning.
\newblock Explainable artificial intelligence.
\newblock {\em DARPA}, 2017.

\bibitem[\protect\citeauthoryear{Guo \bgroup \em et al.\egroup
  }{2021}]{guo2021edge}
W~Guo, X~Wu, et~al.
\newblock Edge: Explaining deep reinforcement learning policies.
\newblock {\em NeurIPS}, 2021.

\bibitem[\protect\citeauthoryear{Hasanbeig \bgroup \em et al.\egroup
  }{2021}]{hasanbeig2021deepsynth}
M~Hasanbeig, NY~Jeppu, et~al.
\newblock Deepsynth: Automata synthesis for automatic task segmentation in deep
  reinforcement learning.
\newblock In {\em AAAI}, 2021.

\bibitem[\protect\citeauthoryear{Hayes and Shah}{2017}]{ImpRoboticController}
B~Hayes and JA~Shah.
\newblock Improving robot controller transparency through autonomous policy
  explanation.
\newblock In {\em HRI}, 2017.

\bibitem[\protect\citeauthoryear{Hein \bgroup \em et al.\egroup
  }{2017}]{HEIN2017fuzzyrl}
D~Hein, A~Hentschel, et~al.
\newblock Particle swarm optimization for generating interpretable fuzzy
  reinforcement learning policies.
\newblock {\em Eng. App. AI}, 2017.

\bibitem[\protect\citeauthoryear{Hein \bgroup \em et al.\egroup
  }{2019}]{genetic}
D~Hein, S~Udluft, and TA~Runkler.
\newblock Interpretable policies for reinforcement learning by genetic
  programming.
\newblock In {\em GeCCO}, 2019.

\bibitem[\protect\citeauthoryear{Heuillet \bgroup \em et al.\egroup
  }{2021}]{alex2020explainability}
A~Heuillet, F~Couthouis, and N~Díaz-Rodríguez.
\newblock Explainability in deep reinforcement learning.
\newblock {\em Knowl. Based Syst.}, 2021.

\bibitem[\protect\citeauthoryear{Huang \bgroup \em et al.\egroup
  }{2018}]{huang2018establishing}
SH~Huang, K~Bhatia, et~al.
\newblock Establishing appropriate trust via critical states.
\newblock In {\em IROS}, 2018.

\bibitem[\protect\citeauthoryear{Huber \bgroup \em et al.\egroup
  }{2019}]{huber2019enhancing}
T~Huber, D~Schiller, and E~Andr{\'e}.
\newblock Enhancing explainability of deep reinforcement learning through
  selective layer-wise relevance propagation.
\newblock In {\em KI}, 2019.

\bibitem[\protect\citeauthoryear{Itaya \bgroup \em et al.\egroup
  }{2021}]{itaya2021visual}
H~Itaya, T~Hirakawa, et~al.
\newblock Visual explanation using attention mechanism in actor-critic-based
  deep reinforcement learning.
\newblock In {\em IJCNN}, 2021.

\bibitem[\protect\citeauthoryear{Iyer \bgroup \em et al.\egroup }{2018}]{katia}
R~Iyer, Y~Li, et~al.
\newblock Transparency and explanation in deep reinforcement learning neural
  networks.
\newblock In {\em AIES}, 2018.

\bibitem[\protect\citeauthoryear{Jhunjhunwala}{2019}]{Jhunjhunwala_policyextraction_2019}
A~Jhunjhunwala.
\newblock Policy extraction via online q-value distillation.
\newblock {\em M. Thesis, Univ. Waterloo}, 2019.

\bibitem[\protect\citeauthoryear{Juozapaitis \bgroup \em et al.\egroup
  }{2019}]{juozapaitis2019explainable}
Z~Juozapaitis, A~Koul, et~al.
\newblock Explainable reinforcement learning via reward decomposition.
\newblock {\em Tech. Report}, 2019.

\bibitem[\protect\citeauthoryear{Koul \bgroup \em et al.\egroup
  }{2019}]{koul2018learning}
A~Koul, S~Greydanus, and A~Fern.
\newblock Learning finite state representations of recurrent policy networks.
\newblock In {\em ICLR}, 2019.

\bibitem[\protect\citeauthoryear{Lage \bgroup \em et al.\egroup
  }{2019}]{lage2019exploring}
I~Lage, D~Lifschitz, et~al.
\newblock Exploring computational user models for agent policy summarization.
\newblock In {\em IJCAI}, 2019.

\bibitem[\protect\citeauthoryear{Landajuela \bgroup \em et al.\egroup
  }{2021}]{landajuela2021discovering}
M~Landajuela, BK~Petersen, et~al.
\newblock Discovering symbolic policies with deep reinforcement learning.
\newblock In {\em ICML}, 2021.

\bibitem[\protect\citeauthoryear{Lin \bgroup \em et al.\egroup
  }{2021}]{lin2021contrastive}
Z~Lin, KH~Lam, and A~Fern.
\newblock Contrastive explanations for reinforcement learning via embedded self
  predictions.
\newblock In {\em ICLR}, 2021.

\bibitem[\protect\citeauthoryear{Lipton}{2018}]{lipton2018mythos}
ZC~Lipton.
\newblock The mythos of model interpretability.
\newblock {\em ACM Queue}, 16(3), 2018.

\bibitem[\protect\citeauthoryear{Liu \bgroup \em et al.\egroup
  }{2018}]{liu2018toward}
G~Liu, O~Schulte, et~al.
\newblock Toward interpretable deep reinforcement learning with linear model
  u-trees.
\newblock In {\em ECML-KDD}, 2018.

\bibitem[\protect\citeauthoryear{Madumal \bgroup \em et al.\egroup
  }{2020}]{madumal2020explainable}
P~Madumal, T~Miller, et~al.
\newblock Explainable reinforcement learning through a causal lens.
\newblock In {\em AAAI}, 2020.

\bibitem[\protect\citeauthoryear{Mnih \bgroup \em et al.\egroup
  }{2013}]{mnih2013playing}
V~Mnih, K~Kavukcuoglu, et~al.
\newblock Playing {A}tari with deep reinforcement learning.
\newblock {\em arXiv preprint arXiv:1312.5602}, 2013.

\bibitem[\protect\citeauthoryear{Molnar}{2019}]{molnar2019}
C~Molnar.
\newblock {\em Interpretable Machine Learning}.
\newblock 2019.

\bibitem[\protect\citeauthoryear{Mott \bgroup \em et al.\egroup
  }{2019}]{mott2019towards}
A~Mott, D~Zoran, et~al.
\newblock Towards interpretable reinforcement learning using attention
  augmented agents.
\newblock In {\em NeurIPS}, 2019.

\bibitem[\protect\citeauthoryear{Murdoch \bgroup \em et al.\egroup
  }{2019}]{murdoch2019definitions}
WJ~Murdoch, C~Singh, et~al.
\newblock Definitions, methods, and applications in interpretable machine
  learning.
\newblock {\em PNAS}, 116(44), 2019.

\bibitem[\protect\citeauthoryear{Olson \bgroup \em et al.\egroup
  }{2021}]{olson2021counterfactual}
ML~Olson, R~Khanna, et~al.
\newblock Counterfactual state explanations for reinforcement learning agents
  via generative deep learning.
\newblock {\em AI}, 2021.

\bibitem[\protect\citeauthoryear{Puiutta and
  Veith}{2020}]{puiutta2020explainable}
E~Puiutta and E~Veith.
\newblock Explainable reinforcement learning: A survey.
\newblock {\em CD-MAKE}, 2020.

\bibitem[\protect\citeauthoryear{Rupprecht \bgroup \em et al.\egroup
  }{2020}]{rupprecht2020finding}
C~Rupprecht, C~Ibrahim, and CJ~Pal.
\newblock Finding and visualizing weaknesses of deep reinforcement learning
  agents.
\newblock In {\em ICLR}, 2020.

\bibitem[\protect\citeauthoryear{Shah \bgroup \em et al.\egroup
  }{2021}]{shah2021minerl}
R~Shah, C~Wild, et~al.
\newblock The {MineRL} basalt competition on learning from human feedback.
\newblock {\em NeurIPS Compet. Track}, 2021.

\bibitem[\protect\citeauthoryear{Shi \bgroup \em et al.\egroup
  }{2020}]{shi2020selfsupervised}
W~Shi, Z~Wang, et~al.
\newblock Self-supervised discovering of causal features: Towards interpretable
  reinforcement learning.
\newblock {\em TPAMI}, 2020.

\bibitem[\protect\citeauthoryear{Silva \bgroup \em et al.\egroup
  }{2020}]{rodriguez2020optimization}
A~Silva, TW~Killian, et~al.
\newblock Optimization methods for interpretable differentiable decision trees
  in reinforcement learning.
\newblock In {\em AISTATS}, 2020.

\bibitem[\protect\citeauthoryear{Sreedharan \bgroup \em et al.\egroup
  }{2020}]{sreedharan2020tldr}
S~Sreedharan, S~Srivastava, and S~Kambhampati.
\newblock Tldr: Policy summarization for factored ssp problems using temporal
  abstractions.
\newblock In {\em ICAPS}, 2020.

\bibitem[\protect\citeauthoryear{Sutton and
  Barto}{2018}]{sutton2018reinforcement}
RS~Sutton and AG~Barto.
\newblock {\em Reinforcement learning: An introduction}.
\newblock MIT press, 2018.

\bibitem[\protect\citeauthoryear{Tang \bgroup \em et al.\egroup
  }{2020}]{tang2020neuroevolution}
Y~Tang, D~Nguyen, and D~Ha.
\newblock Neuroevolution of self-interpretable agents.
\newblock In {\em GeCCO}, 2020.

\bibitem[\protect\citeauthoryear{Topin and Veloso}{2019}]{topin2019policylevel}
N~Topin and M~Veloso.
\newblock Generation of policy-level explanations for reinforcement learning.
\newblock In {\em AAAI}, 2019.

\bibitem[\protect\citeauthoryear{Topin \bgroup \em et al.\egroup
  }{2021}]{topin2021iterative}
N~Topin, S~Milani, et~al.
\newblock Iterative bounding mdps: Learning interpretable policies via
  non-interpretable methods.
\newblock In {\em AAAI}, 2021.

\bibitem[\protect\citeauthoryear{Verma \bgroup \em et al.\egroup
  }{2018}]{verma2018programmatically}
A~Verma, V~Murali, et~al.
\newblock Programmatically interpretable reinforcement learning.
\newblock In {\em ICML}, 2018.

\bibitem[\protect\citeauthoryear{Volodin}{2021}]{volodin2021causeoccam}
S~Volodin.
\newblock Causeoccam: Learning interpretable abstract representations in
  reinforcement learning environments via model sparsity.
\newblock {\em Tech. Report}, 2021.

\bibitem[\protect\citeauthoryear{Wang \bgroup \em et al.\egroup
  }{2019}]{wang2019verbalexplanations}
X~Wang, Y~Schengcheng, et~al.
\newblock Verbal explanations for deep reinforcement learning neural networks
  with attention on extracted features.
\newblock In {\em RO-MAN}, 2019.

\bibitem[\protect\citeauthoryear{Weitkamp \bgroup \em et al.\egroup
  }{2018}]{weitkamp2018visual}
L~Weitkamp, E~van~der Pol, and Z~Akata.
\newblock Visual rationalizations in deep reinforcement learning for {A}tari
  games.
\newblock In {\em BNAIC}, 2018.

\bibitem[\protect\citeauthoryear{Wells and
  Bednarz}{2021}]{wells2021explainable}
L~Wells and T~Bednarz.
\newblock Explainable ai and reinforcement learning—a systematic review of
  current approaches and trends.
\newblock {\em Frontiers in AI}, 2021.

\bibitem[\protect\citeauthoryear{Yau \bgroup \em et al.\egroup
  }{2020}]{yau2020did}
H~Yau, C~Russell, and S~Hadfield.
\newblock What did you think would happen? explaining agent behaviour through
  intended outcomes.
\newblock In {\em NeurIPS}, 2020.

\bibitem[\protect\citeauthoryear{Zahavy \bgroup \em et al.\egroup
  }{2016}]{cluster_drl}
T~Zahavy, N~Ben-Zrihem, and S~Mannor.
\newblock Graying the black box: Understanding dqns.
\newblock In {\em ICML}, 2016.

\bibitem[\protect\citeauthoryear{Zelvelder \bgroup \em et al.\egroup
  }{2021}]{zelvelder2021assessing}
AE~Zelvelder, M~Westberg, and K~Fr{\"a}mling.
\newblock Assessing explainability in reinforcement learning.
\newblock In {\em EXTRAAMAS}, 2021.

\bibitem[\protect\citeauthoryear{Zhang \bgroup \em et al.\egroup
  }{2020}]{zhang2020interpretable}
H~Zhang, A~Zhou, and X~Lin.
\newblock Interpretable policy derivation for reinforcement learning based on
  evolutionary feature synthesis.
\newblock {\em Complex Intell. Syst.}, 2020.

\bibitem[\protect\citeauthoryear{Zhang \bgroup \em et al.\egroup
  }{2021a}]{zhang2021off}
L~Zhang, X~Li, et~al.
\newblock Off-policy differentiable logic reinforcement learning.
\newblock In {\em ECML-KDD}, 2021.

\bibitem[\protect\citeauthoryear{Zhang \bgroup \em et al.\egroup
  }{2021b}]{zhang2021learning}
Q~Zhang, X~Ma, et~al.
\newblock Learning to discover task-relevant features for interpretable
  reinforcement learning.
\newblock {\em I-RAL}, 2021.

\end{thebibliography}

\end{document}